\documentclass{article}

\usepackage[final]{neurips_2020}
\usepackage[utf8]{inputenc} % allow utf-8 input
\usepackage[T1]{fontenc}    % use 8-bit T1 fonts
\usepackage{hyperref}       % hyperlinks
\usepackage{url}            % simple URL typesetting
\usepackage{booktabs}       % professional-quality tables
\usepackage{amsfonts}       % blackboard math symbols
\usepackage{nicefrac}       % compact symbols for 1/2, etc.
\usepackage{microtype}      % microtypography
\usepackage{CJKutf8}
\usepackage{enumerate}
\usepackage{comment}
\usepackage{multirow}
\usepackage{amsmath}
\usepackage{makecell}
\usepackage{graphicx}
\usepackage{booktabs}
\usepackage{multirow}
\usepackage{array}
\usepackage{comment}
\usepackage{multirow}
\usepackage{mathrsfs}

\title{EpilepsyLLM: Domain-Specific Large Language Model Fine-tuned with Epilepsy Medical Knowledge}

\author{
Xuyang Zhao$^{1,2}$, Qibin Zhao$^{2}$ and Toshihisa Tanaka$^{1}$\\ \\
1. Department of Electrical Engineering and Computer Science,\\
Tokyo University of Agriculture and Technology, Tokyo, Japan\\
2. Tensor Learning Team,\\
RIKEN Center for Advanced Intelligence Project, Tokyo, Japan\\
}

\begin{document}

\maketitle

% % % % % % % % % % % % % % % % % % % % % % % % % % % % % %
% % % % % % % % % % % % % % % % % % % % % % % % % % % % % %
\begin{abstract}
With large training datasets and massive amounts of computing sources, large language models (LLMs) achieve remarkable performance in comprehensive and generative ability. Based on those powerful LLMs, the model fine-tuned with domain-specific datasets  posseses more specialized knowledge and thus is more practical like medical LLMs. However, the existing fine-tuned medical LLMs are limited to general medical knowledge with English language. For disease-specific problems, the model's response is inaccurate and sometimes even completely irrelevant, especially when using a language other than English. In this work, we focus on the particular disease of Epilepsy with Japanese language and introduce a customized LLM termed as EpilepsyLLM. Our model is trained from  the pre-trained LLM by fine-tuning technique using datasets from the epilepsy domain. The datasets contain knowledge of basic information about disease, common treatment methods and drugs, and important notes in life and work. The experimental results demonstrate that EpilepsyLLM can provide more reliable and specialized medical knowledge responses.
\end{abstract}

% % % % % % % % % % % % % % % % % % % % % % % % % % % % % %
% % % % % % % % % % % % % % % % % % % % % % % % % % % % % %
\section{Introduction}
In recent years,  large language models (LLMs) have demonstrated impressive performance, especially success in instruction understanding and human-like response generation. In the natural language processing (NLP) benchmark, LLMs continuously refreshed records. Most LLMs are built based on Transformer structure~\citet{vaswani2017attention}, and with the help of large training data and massive computing resource.
OpenAI has released a series models of GPT-3~\citet{brown2020language}, InstructGPT~\citet{ouyang2022training} and GPT-4~\citet{openai2023gpt4}, and the performance of the models is constantly updated. However, the details of the model such as architecture, training strategies, training data and other information have not been released.
LLaMA~\citet{touvron2023llama} is released by the Meta which is open-source in the research field, containing four different parameter models of 7, 13, 30 and 65 billons. At the same time, the model also achieved good performance on most benchmarks. LLaMA (13B) outperforms GPT-3 and  is ten times smaller.
For the largest 65B parameter model, the performance is similar to some well-known models such as Chinchilla and PaLM-540B.

In order to further improve the performance of the LLMs, Stanford Alpaca~\citet{alpaca} has proposed an efficient and low-cost fine-tuning method. The Alpaca method just uses 52K instruction-following demonstrations to fine-tune the pre-trained LLaMA (7B) model and the Alpaca behaves qualitatively similarly to OpenAI’s GPT-3.5 (text-davinci-003).
The instruction-following demonstrations are built from 175 human-written instruction-output pairs and use GPT-3.5 (text-davinci-003) to generate more instructions.
In this mode, the performance of the basic model (LLaMA 7B) is efficient (8 80GB A100s with 3 hours) and low cost (<600\$) improved to the GPT-3.5 (text-davinci-003) level.

Usually,  LLMs focus on general tasks, and while these LLMs achieve amazing performance, in subdivided professional fields, the performance of the LLMs is often not professional and accurate enough.
In order to improve the professional performance of LLMs in a specific field, it is often possible to fine-tune the model using specific field knowledge.
In the medical field, there have been some works such as PubMedBERT~\citet{gu2021domain}, BioLinkBert~\citet{yasunaga-etal-2022-linkbert}, BioMedLM~\citet{venigalla2022biomedlm}, BioGPT~\citet{luo2022biogpt}, Med-PLaM~\citet{tu2023generalist}, and ChatDoctor~\citet{yunxiang2023chatdoctor}.
Among them, the medical knowledge includes patient-physician conversations, PubMed abstracts \& PubMed Central full text articles, and e-identified clinical notes.
With the help of medical knowledge, fine-tuned LLMs achieves better performance on medical tasks. In the United States Medical Licensing Examination (USMLE), MedPaLM 2 achieved the highest score of $86.5$.

LLMs fine-tuned by the medical knowledge achieve better performance  in medical general tasks in  English language. Unlike the common medical LLMs, this work presents a model of EpilepsyLLM that  focuses on a particular disease of epilepsy  and uses  Japanese language.
When using LLMs in the medical field, it is of more importance that the professionalism of the model output results. Our  dataset for fine-tuning is professional knowledge about epilepsy collected from the Internet and is prepared as instruction-following demonstrations. Furthermore, the fine-tuning dataset uses the Japanese language.
We adopt the pre-trained LLaMA (7B) and LLM-JP~\footnote{https://github.com/llm-jp} as the base model. The experimental results show that  LLMs fine-tuned with epilepsy knowledge achieves the highest performance in testing. Compared with LLaMA and LLM-JP for general tasks or medical LLMs that have been fine-tuned with medical knowledge,  EpilepsyLLM shows more professional and reliable answers when faced with epilepsy knowledge.
The experimental results also confirmed that by using more specific domain knowledge to fine-tune the LLMs, the performance of the model in this domain can be greatly improved.

The rest of the article is organized as follows. Section 2 describes the fine-tuning dataset and EpilepsyLLM model.  Section 3 explains the experimental results. The discussion and conclusion are presented in Sections 4 and 5, respectively.

% % % % % % % % % % % % % % % % % % % % % % % % % % % % % %
% % % % % % % % % % % % % % % % % % % % % % % % % % % % % %
\section{Approach}
In this work, we use epilepsy knowledge to fine-tune the LLMs to improve the epilepsy-specific domain performance. The model shows the more reliable medical knowledge responses and better support in Japanese.

% % % % % % % % % % % % % % % % % % % % % % % % % % % % % %
\subsection{Epilepsy Knowledge}
Epilepsy is a common brain neurological disorder disease that affects 50 million people around the world World Health Organization (WHO)~\citet{world2019epilepsy}.
To improve patients’ quality of life, there are some institutions that provide basic knowledge about epilepsy, including basic information about the disease, treatment methods, life precautions, school work, etc.
In this work, the epilepsy dataset is collected from the Japan Epilepsy Association~\footnote{https://www.jea-net.jp/}, Epilepsy Information Center~\footnote{https://shizuokamind.hosp.go.jp/epilepsy-info/} and Tenkan Net~\footnote{https://www.alfresa-pharma.co.jp/general/tenkan/faq/index.html}.
Referring to Alpaca~\citet{alpaca}, the dataset is prepared as instruction-following demonstrations, some examples in the dataset are shown in Table~\ref{table_dataset}.
There are more than two hundred samples used for fine-tuning the model and 24 samples used for model testing.

% % % % % % % % % % % % % % % % % % % % % % % % % % % % % %
\subsection{Base Model}
Two open-source models of LLaMA and LLM-JP are used as base models.
LLaMA~\citet{touvron2023llama} is a collection of foundation language models ranging from 7B to 65B parameters (four different parameters of 7, 13, 30, and 65 billion).
The LLaMA is trained on trillions of tokens, and all the training data is publicly available. The training data used in LLaMA in shown in Table~\ref{table_training_data}.
As a medium-size model, the LLaMA (13B) reaches the performance of GPT-3.
By using Alpaca fine-tuning method, the LLaMA (7B) behaves qualitatively similarly to OpenAI’s GPT-3.5 (text-davinci-003).

Another base model is LLM-JP, which is trained by Japanese and English language data.
LLM-JP has two different model size with the number of parameters of 1.3 and 13 billion.
In multiple Japanese benchmark, LLM-JP achieved promising performance.

\begin{table}[htbp]
  \caption{Examples of epilepsy dataset.}
  \label{table_dataset}
  \scriptsize
  \centering
  
  \begin{tabular}{| p{1cm} | p{12cm} |}
    \toprule
    Question & Answer \\ \midrule
    
    \begin{CJK}{UTF8}{ipxm}てんかんはどんな病気ですか\end{CJK} & 
    \begin{CJK}{UTF8}{ipxm}てんかん発作は大脳神経細胞の過剰な亢奮（てんかん発射）によって生じます。てんかん発作を繰り返し、慢性に経過する大脳の病気がてんかんです。ただ1  回だけの発作や高熱のときだけに起こる発作はてんかんとは区別されます。また、脳炎の急性期や頭部外傷の直後に起こったけいれん発作も、それだけであれば、てんかんとは区別されます。慢性に経過す  ることがてんかんの特徴です。てんかん発作は大脳の機能が種々の形で表出されるので、発作症状はきわめ て多彩です。しかし、ひとりの患者の発作症状はほぼ一定しています。なぜてんかんになるのか、なぜてんかん発作が起こるかは十分にはわかっていません。大脳の神経細胞ネットワークの亢奮と抑制の微妙なバランスの破綻に求める学説が有力です。てんかん発作を起こしやすい下  地（素因）と大脳の機能を不安定にする脳損傷（外因）とが複雑に絡み合っててんかんが発病すると推測されています。\end{CJK} \\ \midrule
    
    \begin{CJK}{UTF8}{ipxm}てんかん発作が起こることで、脳にダメージを受けますか\end{CJK} &
    \begin{CJK}{UTF8}{ipxm}てんかん重積状態という特別な状態をのぞけば、通常、一度の発作で脳がダメージを受け ることはありません。てんかん重積状態とは、30 分以上発作がとまらなかったり、ひとつの発作が終わった後、意識が完全に回復する前に次の発作が起こることを  30  分以上繰り返す状態を指し、注射など薬物で発作をとめる必要のある緊急事態です。ただし、長い経過を見ますと、記憶障害、知能低下や行動障害、精神医学的問題を合併することがありま  す。これらは発作を繰り返した影響や、過ぎた薬物の服用、心理的要因、てんかんの原因となっているもと の病気の影響など種々の要因が関与しています。てんかんが難治に経過する場合は多種類の抗てんかん薬による治療はできるだけ避け、場合によってはて んかん外科手術を早めに検討したほうがいいことがあります。また、てんかんではありませんが、乳幼児期に熱性けいれんなどのけいれんが長時間持続すると、側頭葉  の内側にある海馬などの萎縮が起こり、それが数年から 10 年後に側頭葉てんかんの原因になることが知られています。\end{CJK} \\ \midrule
    
    \begin{CJK}{UTF8}{ipxm}高齢で発病するてんかんの特徴を教えてください\end{CJK} & 
    \begin{CJK}{UTF8}{ipxm}てんかんの多くは小児期に発病します。2008 年2 月～ 11 月に当院（静岡てんかん・神経医療センター）でてんかんと診断された928 名のうち、751 名（80.9 ％）が20 歳までに発病していました。20 歳を過ぎるとてんかんの発病率は次第に低下していきますが、60 歳を越えると再び増加に転ずるといわれています。脳血管障害など、高齢になって新たに生ずる脳の器質的な障害を背  景に、てんかんが発病しやすくなると考えられています。当院の統計でも高齢になるとわずかに発病率が増え、15 名（1.6 ％）が60 歳以降に発病していました。推定病因は脳梗塞2、脳膿瘍2、脳腫瘍 1、脳外傷1、長年の大酒1 名で、他は不明でした。このように 器質性の病因が約半数に認められるのは、てんかん全体から見れば多い数字です。一方、熱性けいれんの既往や、てんかんまたは熱性けいれんの家族歴をもつ例は1  名もありませんでした。診断は15 名全例が症候性部分てんかんで、うち4 名は側頭葉に焦点を認めました。発作型は全般性強直間代発作が 10、複雑部分発作が 7、単純部分発作が5、非けいれん性てんかん重積状態の疑いが2 名でした（重複あり）。発作頻度は年単位 5、月単位 7、週単位 2、日単位1 名でした。ほぼ全例が結婚してお子さんもあり、お仕事もしてこられた方々です。一般に、高齢発病のてんかんは薬物治療で比較的容易に発作がコントロールされることが多いといわれています。高齢者では血中濃度が上昇しやすいので、服用量を決定する際には注意が必要です。\end{CJK}\\ \midrule
    
    \begin{CJK}{UTF8}{ipxm}外科治療は誰でも受けられますか\end{CJK} & 
    \begin{CJK}{UTF8}{ipxm}手術という侵襲を考慮すると、外科治療の対象は、薬物治療で発作が抑制されない難治症例に限られます。薬が効かないと判定するには、1）正しく診断されたてんかんに対して、2）適切な薬が2 種類以上使われたにもかかわらず、3）2 年以上発作が抑制されていない状態がめやすとなります。ただし、子どもの場合、病状によっては発育への影響を考慮して、2  年を待たずに積極的に外科治療を検討することもあります。根治手術は、切除可能な病変が原因で、大脳の一部から発作が始まるてんかんを対象とします。一方、緩 和手術は、根治手術の対象とはならないが、転倒する発作で外傷が絶えない症例などに検討されます。手術適応を検討する際に最も大切なことは、術前の病状だけではなく、術後の生活への期待まで視野に入 れて、手術の意義について患者さんやそのご家族とよく話し合い、理解を共有することです。\end{CJK}\\ \midrule
    
    \begin{CJK}{UTF8}{ipxm}発作が起きると、子どもの成長に影響がありますか\end{CJK} & 
    \begin{CJK}{UTF8}{ipxm}てんかんのある患者さんの身長や体重の変化をさす「成長」に関しての研究は、認知・知 能・運動などの「発達」に関する研究に比べ多くなく、はっきりしたことはわかっていません。両者ともにさまざまな因子に影響されると思われます。「発達」についていえば、症候性ウエスト症候群、レンノックス症候群などの症候性全般てんかんは、て  んかんの発病とともに知的退行（それまでできていたことが徐々にできなくなる）を起こす悪性のてんかん  です。いわばてんかんの緊急事態であり、早急な診断と治療が知能の保護につながります。てんかんの種類、基礎疾患の有無や種類、てんかん発作が抑制されているかどうかによっても、発達への影響は異なります。また、抗てんかん薬の種類によっては鎮静効果の強いものや落ち着きをなくすなどの行動や認知に影響するもの、食欲に影響を与えるものもありますし、単剤か多剤併用かによっても違います。  これらは患者さん一人ひとり違っていますから、詳しいことは主治医あるいはてんかん専門の先生に尋ねてみてください。\end{CJK}\\
    \bottomrule
    
  \end{tabular}
\end{table}

\begin{table}[htbp]
  \caption{Training and fine-tuning dataset.}
  \label{table_training_data}
  \scriptsize
  
  \centering
  \begin{tabular}{l | l | l} \toprule
    Dataset & Content & Language     \\ \midrule
    \multirow{7}*{LLaMA Training} & CommonCrawl    & English      \\
                    & C4             & English      \\
                    & Github         & English      \\
                    & Wikipedia      & 20 languages \\
                    & Books          & English      \\
                    & ArXiv          & English      \\
                    & Stack Exchange & English      \\ \midrule

    Alpaca          & Instruction-following & English \\ \midrule
    Japanese Alpaca & Instruction-following & Japanese (Translate from Alpaca) \\ \midrule \midrule
    
    \multirow{5}*{LLM-JP Training} & mC4          & Japanese \\
                                   & Wikipedia    & Japanese \\
                                   & Pile         & English \\
                                   & Wikipedia    & English \\
                                   & Stack (code) & English \\ \midrule
                                   
    Jaster         & Instruction-following & Japanese \\ \midrule
    Japanese Dolly & Instruction-following & Japanese (Translate from Dolly) \\ \midrule 
    Japanese Oasst & Instruction-following & Japanese (Translate from Oasst) \\ \midrule \midrule

    Epilepsy Dataset & Epilepsy knowledge    & Japanese \\ \bottomrule

  \end{tabular}
\end{table}

% % % % % % % % % % % % % % % % % % % % % % % % % % % % % %
% % % % % % % % % % % % % % % % % % % % % % % % % % % % % %
\section{Experiments}
In this section, we evaluate the performance of the proposed method through a series of experiments. Since we have limited computing resources (4 80GB A100s), for LLaMA experiments, LLaMA (7B) is used as base model for fine-tuning and the larger models of LLaMA (13B) and LLaMA (30B) are used directly for inference. The Alpaca dataset is also used for fine-tuning.
Since the epilepsy dataset is collected in Japanese, to verify the impact of language on LLMs performance, a translated version of the Alpaca dataset is also used for fine-tuning.
The Alpaca is translated using ChatGPT, the Japanese-Alpaca data is provided from Github~\footnote{https://github.com/masa3141/japanese-alpaca-lora/tree/main}.
For the LLM-JP experiments, the LLM-JP (1.3B) is used as base model for fine-tuning, and the LLM-JP (13B) is used directly for inference.
Three different fine-tuned models of
LLM-jp-13B-instruct-full-jaster, LLM-jp-13B-instruct-full-jaster-dolly-oasst and LLM-jp-13B-instruct-full-dolly-oasst
% (jaster, dolly-oasst and jaster-dolly-oasst) 
provided by LLM-JP are also used for inferences.

In the evaluation, four metrics of BLEU, METEOR, ROUGE-L, and SPICE are used to evaluate the performance of the LLMs. 
The experimental results are shown in Table~\ref{table_result}.
By fine-tuning the base LLMs using epilepsy knowledge data, the model's performance is greatly improved.
Although the LLM-JP (1.3B) has the fewest parameters, it achieves the highest performance in the epilepsy task due to the large amount of Japanese text included in the training data.

%Six comparative experiments will be performed.
%\begin{enumerate}[1)]
%  \item Pre-trained LLaMA.
%  \item Fine-tuning Pre-trained LLaMA with Epilepsy data.
%  \item Fine-tuning Pre-trained LLaMA with Alpaca data.
%  \item Fine-tuning 3) Model with Epilepsy data.
%  \item Fine-tuning Pre-trained LLaMA with Japanese data.
%  \item Fine-tuning 5) Model with Epilepsy data.
%\end{enumerate}

%\begin{table}
%  \caption{Sample table title}
%  \label{sample-table}
%  \small
%
%  \centering
%  \begin{tabular}{c | c | c | c c c c c c c c}
%    \toprule
%    % \multicolumn{2}{c}{Part}                   \\
%    % \cmidrule(r){1-2}
%    Model & FT 1          & FT 2          & BLEU-1 & BLEU-2 & BLEU-3 & BLEU-4 & METEOR & ROUGE-L & CIDEr  & SPICE  \\ \midrule
%    LLaMA & --            & --            & 0.0173 & 0.0007 & 0      & 0      & 0.0237 & 0.0234  & 0      & 0.0069 \\
%    LLaMA & Epilepsy      & --            & 0.2196 & 0.0202 & 0.0099 & 0.0018 & 0.1877 & 0.2709  & 0      & 0.1098 \\ \midrule
%    LLaMA & Alpaca        & --            & 0.0108 & 0      & 0      & 0      & 0.0185 & 0.0382  & 0      & 0.0186 \\
%    LLaMA & Alpaca        & Epilepsy      & 0.1690 & 0.0314 & 0.0138 & 0      & 0.1469 & 0.2417  & 0      & 0.1347 \\ \midrule
%    LLaMA & Japanese      & --            & 0.1438 & 0.0188 & 0.0102 & 0      & 0.1111 & 0.2175  & 0      & 0.1080 \\
%    LLaMA & Japanese      & Epilepsy      & 0.1939 & 0.0155 & 0.0068 & 0      & 0.1717 & 0.2700  & 0      & 0.1262 \\
%    \bottomrule
%  \end{tabular}
%\end{table}

\begin{table}[htbp]
  \caption{LLMs performance on epilepsy tasks.}
  \label{table_result}
  \begin{center}
  \scriptsize % \small

  \begin{tabular}{l | l | l | c c c c | c} \toprule
    Model & Fine-tuning 1 & Fine-tuning 2 & BLEU & METEOR & ROUGE-L & SPICE & Mean \\ \midrule
    
    BioMedLM & --  & -- & 0.0058 & 0.0091 & 0.0092 & 0.0000 & 0.0060 \\ \midrule \midrule
    % BioMedLM & Epilepsy Q\&A & --           &        &        &        &        &  \\ 

    LLaMA (7B) & --              & --            & 0.0173 & 0.0237 & 0.0234 & 0.0069 & 0.0178 \\
    LLaMA (7B) & Epilepsy Data   & --            & \pmb{0.2256} & \pmb{0.1836} & \pmb{0.2820}  & 0.1045 & \pmb{0.1989} \\ 

    LLaMA (7B) & Alpaca          & --            & 0.0273 & 0.0418 & 0.0639 & 0.0439 & 0.0442 \\
    LLaMA (7B) & Alpaca          & Epilepsy Data & 0.1701 & 0.1705 & 0.2347 & 0.1070 & 0.1706 \\ 

    LLaMA (7B) & Japanese Alpaca & --            & 0.1637 & 0.1380 & 0.2217 & 0.0876 & 0.1528 \\
    LLaMA (7B) & Japanese Alpaca & Epilepsy Data & 0.2037 & 0.1678 & 0.2668 & \pmb{0.1308} & 0.1923 \\ 

    LLaMA (13B)& --              & --            & 0.0281 & 0.0559 & 0.0417 & 0.0057 & 0.0328 \\
    LLaMA (30B)& --              & --            & 0.0281 & 0.0572 & 0.0417 & 0.0057 & 0.0332 \\ \midrule \midrule

    LLM-JP (1.3B) & --            & --           & 0.1418 & 0.1793 & 0.1805 & 0.0144 & 0.1290 \\ 
    LLM-JP (1.3B) & Epilepsy Data & --           & \pmb{0.2351} & \pmb{0.2314} & \pmb{0.2631} & \pmb{0.0727} & \pmb{0.2006} \\ 

    LLM-JP (13B)  & --            & --           & 0.1673 & 0.2102 & 0.2010 & 0.0198 & 0.1496 \\ 
    % LLM-JP (13B)  & Epilepsy Q\&A & --            & 0.1142 & 0.1372 & 0.1943 & 0.0444 & 0.1225 \\ 

    LLM-JP (13B)  & Jaster                   & --            & 0.0004 & 0.0192 & 0.0174 & 0.0160 & 0.0132 \\ 
    % LLM-JP (13B)  & Jaster                   & Epilepsy Q\&A &        &        &        &        &        \\ \midrule
    LLM-JP (13B)  & Japanese Dolly           & --            & 0.0880 & 0.0891 & 0.1421 & 0.0647 & 0.0960 \\ 
    % LLM-JP (13B)  & Japanese Dolly           & Epilepsy Q\&A &        &        &        &        &        \\ \midrule
    LLM-JP (13B)  & Jaster \& Japanese Dolly & --            & 0.0712 & 0.0889 & 0.1295 & 0.0712 & 0.0902 \\ \bottomrule
    % LLM-JP (13B)  & Jaster \& Japanese Dolly & Epilepsy Q\&A &        &        &        &        &        \\
  \end{tabular}
  
  \end{center}
\end{table}

% % % % % % % % % % % % % % % % % % % % % % % % % % % % % %
% % % % % % % % % % % % % % % % % % % % % % % % % % % % % %
\section{Discussion}
Although the LLMs exhibit strong performance on general tasks. However, in some professional domains, the model's response sometimes lacks professionalism and is not accurate enough.
In order to make the response generated by the LLMs more professional and reliable, some works use the knowledge of specific domains to fine-tune the pre-trained model. In this work, we try to use more granular epilepsy knowledge to fine-tune the pre-trained model using non-English languages.

In the experiments, two different open-source LLMs are used as base models, and the fine-tuning data is collected from public websites. From the experimental results, the Japanese model of LLM-JP (1.3B) achieved the highest performance after being fine-tuned with epilepsy knowledge. Due to the lack of Japanese language support, the medical model of BioMedLM also do not show strong performance.

For the LLaMA that has not been fine-tuned, more parameters (from 7B to 30B) do not bring about a significant performance improvement. Due to the lack of professional epilepsy medical knowledge in the model's training data, and the very low proportion of Japanese data, for Japanese epilepsy test questions, the model lacks the ability to understand and respond to the Epilepsy questions. For LLaMA (7B), the performance of the model is significantly improved by fine-tuning using Japanese epilepsy data. 
% The Alpaca data effectively improved the LLaMA performance in general tasks, but in the epilepsy domain, the fine-tuning reduces the model performance.
The Alpaca data effectively improved the LLaMA performance in general tasks~\citet{alpaca}, in the epilepsy task, the model performance has also been slightly improved.
However, based on the alpaca model, using the second fine-tuning of epilepsy data, the model performance did not reach the strongest performance.
% the final performance of the model is not improved through two fine-tuning of Alpaca data and epilepsy data.
By using the translated Japanese Alpaca and the epilepsy data, the performance of the two-times fine-tuned model is improved.

Since the LLM-JP uses a large amount of Japanese data in the training process, the model shows a strong performance on Japanese tasks.
The performance of LLM-JP (1.3B) fine-tuned based on epilepsy data has been greatly improved from $0.129$ to $0.2006$.
For tasks in specific domains, using specific data for fine-tuning can achieve significant performance improvements.

% % % % % % % % % % % % % % % % % % % % % % % % % % % % % %
% % % % % % % % % % % % % % % % % % % % % % % % % % % % % %
\section{Conclusion}
In this work, to improve the professionalism and reliability of LLMs in epilepsy tasks, we try to use epilepsy knowledge to fine-tune the model.
Different from common medical LLMs, we use the more specific disease knowledge to fine-tune the model. 
From the experimental results, by narrowing the scope of the domain, smaller data can also improve the performance of the model in a specific domain.

% % % % % % % % % % % % % % % % % % % % % % % % % % % % % %
% % % % % % % % % % % % % % % % % % % % % % % % % % % % % %
\section*{Statement}
EpilepsyLLM is dedicated to the research of LLMs in the medical field. The medical knowledge used in model training and testing is obtained from publicly accessible websites. The response content generated by the model cannot be guaranteed and cannot be used as a substitute for professional medical treatment.

% % % % % % % % % % % % % % % % % % % % % % % % % % % % % %
% % % % % % % % % % % % % % % % % % % % % % % % % % % % % %
%\section*{References}
%References follow the acknowledgments. Use unnumbered first-level heading for the references.
%\medskip

%\small
%
%[1] Alexander, J.A.\ \& Mozer, M.C.\ (1995) Template-based algorithms for connectionist rule extraction. In G.\ Tesauro, D.S.\ Touretzky and T.K.\ Leen (eds.), {\it Advances in Neural Information Processing Systems 7}, pp.\ 609--616. Cambridge, MA: MIT Press.
%
%[2] Bower, J.M.\ \& Beeman, D.\ (1995) {\it The Book of GENESIS: Exploring Realistic Neural Models with the GEneral NEural SImulation System.}  New York: TELOS/Springer--Verlag.
%
%[3] Hasselmo, M.E., Schnell, E.\ \& Barkai, E.\ (1995) Dynamics of learning and recall at excitatory recurrent synapses and cholinergic modulation in rat hippocampal region CA3. {\it Journal of Neuroscience} {\bf 15}(7):5249-5262.

\clearpage
\bibliographystyle{plainnat}
\bibliography{neurips_2020}

\end{document}